\documentclass[conference]{IEEEtran}
\IEEEoverridecommandlockouts
\usepackage{cite}
\usepackage{amsmath,amssymb,amsfonts}
\usepackage{algorithmic}
\usepackage{graphicx}
\usepackage{textcomp}
\usepackage{xcolor}
\usepackage{tabularx}  
\usepackage{booktabs}  
\usepackage{url} 
\usepackage{hyperref}
\usepackage{placeins}

\def\BibTeX{{\rm B\kern-.05em{\sc i\kern-.025em b}\kern-.08em
    T\kern-.1667em\lower.7ex\hbox{E}\kern-.125emX}}
\begin{document}

\title{Leveraging Multimodal Models for Enhanced Neuroimaging Diagnostics in Alzheimer's Disease\\

}

\author{
\IEEEauthorblockN{Francesco Chiumento}
\IEEEauthorblockA{\textit{School of Electronic Engineering} \\
\textit{Dublin City University}\\
Glasnevin, Dublin 9, Ireland \\
francesco.chiumento2@mail.dcu.ie}
\and
\IEEEauthorblockN{Mingming Liu}
\IEEEauthorblockA{\textit{Insight Research Ireland Centre for Data Analytics} \\
\textit{Dublin City University}\\
Glasnevin, Dublin 9, Ireland \\
mingming.liu@dcu.ie}
}

\maketitle

\begin{abstract}
The rapid advancements in Large Language Models (LLMs) and Vision-Language Models (VLMs) have shown great potential in medical diagnostics, particularly in radiology, where datasets such as X-rays are paired with human-generated diagnostic reports. However, a significant research gap exists in the neuroimaging field, especially for conditions such as Alzheimer’s disease, due to the lack of comprehensive diagnostic reports that can be utilized for model fine-tuning. This paper addresses this gap by generating synthetic diagnostic reports using GPT-4o-mini on structured data from the OASIS-4 dataset, which comprises 663 patients. Using the synthetic reports as ground truth for training and validation, we then generated neurological reports directly from the images in the dataset leveraging the pre-trained BiomedCLIP and T5 models. Our proposed method achieved a BLEU-4 score of 0.1827, ROUGE-L score of 0.3719, and METEOR score of 0.4163, revealing its potential in generating clinically relevant and accurate diagnostic reports.
\end{abstract}


\section{Introduction}
\label{sec:intro}
Degenerative diseases are conditions that gradually damage and destroy parts of the cells of the nervous system, particularly in areas such as the brain. These diseases typically develop slowly, and the effects and symptoms generally manifest in the later stages.
Among the most common degenerative diseases is Alzheimer's, which is estimated to currently affect 6.9 million Americans aged 65 or older, and remains the fifth-leading cause of death \cite{2024AlzheimerDisease2024}.
Alzheimer's disease (AD) develops in different stages, and is the most common cause of dementia, accounting for 60-80\% of all cases \cite{duongDementia2017}. Currently, there is no definitive cure; however, early diagnosis of this condition can lead to a slowdown in its progression and an improvement in the patient's quality of life. To meet the need for fast and accurate diagnoses, there is an increasing reliance on automatic diagnostic systems based on machine learning methods. Diagnostic reports provide essential textual descriptions and are important for the early diagnosis and treatment of the disease. The interpretation of these reports can indeed influence patient outcomes. However, interpreting biomedical images to generate diagnostic reports can take a considerable amount of time, even for the most experienced clinicians \cite{2024AlzheimerDisease2024,kumarAlzheimerDisease2024, tuGeneralistBiomedicalAI2023}. 

Deep learning methods, particularly those based on Convolutional Neural Networks (CNNs) and transformer architectures, have shown great potential in the detection of Alzheimer's disease, thanks to their ability to learn complex patterns and representations from large-scale datasets \cite{alsubaieAlzheimersDiseaseDetection2024}. However, the use of CNNs presents several limitations, such as the inability to capture long-term dependencies and the absence of an attention mechanism. Furthermore, these models are frequently criticized for their lack of interpretability. The hierarchical and non-linear nature of their processing can make it challenging to understand how they arrive at a particular decision \cite{shahComputeraidedDiagnosisAlzheimers2024}. Additionally, one of the main challenges is the efficient integration of medical images with structured data. Despite the potential benefits of combining these data, their integration can affect the quality of classification performance. To address these limitations, recent research has explored the potential of more advanced language models and multimodal approaches. These models can process and integrate various data types, showing promising results in terms of accuracy and interpretability for Alzheimer’s disease \cite{jiVisionLanguageModelGenerating2024}. Transformers leverage their attention mechanisms to capture global contextual information, which can make them more interpretable in decision-making processes particularly for tasks that require understanding long-range dependencies or a comprehensive view of the context \cite{FocusedAttentionTransformers2022,mauricioComparingVisionTransformers2023,chaudhari2021attentive}.

To the best of our knowledge, the generation of synthetic diagnostic reports for Alzheimer's disease using multimodal approaches has not been previously explored in the literature. Specifically, our research question in this work is to determine how synthetic diagnostic reports can bridge the gap between existing neuroimaging datasets and the training requirements of Visual Language Models (VLMs) and Large Language Models (LLMs) for Alzheimer's diagnosis. With this in mind, the main contributions of our work are as follows:

\begin{itemize}
    \item We generate synthetic diagnostic reports to address the lack of textual data in neuroimaging, facilitating the fine-tuning of multimodal models for Alzheimer's diagnosis;
    \item We propose a framework using BiomedCLIP and T5 to combine visual features from MR images with clinical descriptions, extending multimodal model applications to neuroimaging datasets with a particular focus on the OASIS;
    \item We integrate MR images and clinical data to analyze both visual and non-visual information, capturing relationships between brain morphology and cognitive decline to improve diagnostic accuracy;
    \item Finally, we evaluate the quality of the generated reports using BLEU, ROUGE, and METEOR metrics on the OASIS-4 dataset, using synthetic reports as ground truth.
\end{itemize}

The structure of the paper is as follows. Section \ref{sec:related work} provides an overview of the recent use of LLMs and VLMs for generating diagnostic reports based on biomedical images, emphasizing the challenges in neuroimages. The OASIS cohorts are presented in Section \ref{sec:datasets}, with a particular focus on the main cohort: OASIS-4. Section \ref{sec:methods} provides an overview of the models, techniques, and overall approach used in the study. Section \ref{sec:results} presents the results obtained from the models, analyzing their performance and the quality of the generated reports, while highlighting strengths and areas for improvement. Finally, Section \ref{sec:conclusion} concludes the study, discusses the model’s current limitations, and outlines some directions for future research.

\section{Related Work}
\label{sec:related work}
\subsection{Current Advancements in LLMs and VLMs}
LLMs and VLMs, trained on vast datasets, have recently shown incredible performance in diagnostic report generation, significantly enhancing accuracy and consistency \cite{vanLargeVisualLanguage2024,jiVisionLanguageModelGenerating2024}.
The development of foundation models has been crucial, impacting various applications in language processing and computer vision. Foundation models such as CLIP (Contrastive Language-Image Pretraining) and PaLM are particularly notable for their ability to be adapted to various tasks without requiring significant modifications to their parameters \cite{tuGeneralistBiomedicalAI2023}. These models utilize enormous amounts of data and computational resources during training; this enables them to generalize effectively across various domains. The integration of multimodal data, such as text and images, has been particularly effective, allowing these models to generate coherent text across different modalities. For example, models like CLIP create visual representations associated with linguistic descriptions, making it possible to perform complex image-text alignment tasks \cite{radfordLearningTransferableVisual2021}.

In the biomedical domain, several models that employ these advancements have been developed. For example, BiomedCLIP \cite{zhangBiomedCLIPMultimodalBiomedical2024} combines textual and visual information to map medical images and text into a common representation space. Based on the CLIP architecture, BiomedCLIP employs two separate encoders: a PubMedBERT-based textual encoder and an advanced version of a Vision Transformer (ViT) pretrained on the PMC-15M dataset, a large collection of 15 million image-text pairs extracted from scientific articles in PubMed Central (PMC). This multimodal approach allows BiomedCLIP to perform well in the alignment of image-text pairs specifically in the medical field. In addition, Med-PaLM 2 has shown excellent results in extracting relevant clinical information and generating detailed medical reports, achieving 86.5\% accuracy on the MedQA dataset, a benchmark based on the USMLE style dataset \cite{vanLargeVisualLanguage2024,tuGeneralistBiomedicalAI2023}. Despite these advancements, LLM-based chatbots often generate responses that, while promising, are not reliable enough for real-world clinical settings, highlighting the need for further refinement \cite{vanLargeVisualLanguage2024}. One of the most prominent LLMs explored in this study is the T5 model (Text-to-Text Transfer Transformer) \cite{raffelExploringLimitsTransfer2019,huggingfaceT5}. T5 is a unified framework for natural language processing tasks, treating every NLP problem as a text-to-text task. It utilizes a modified encoder-decoder transformer, and its architecture was trained on the\textit{ Colossal Clean Crawled Corpus (C4)} \cite{dodgeDocumentingLargeWebtext2021} dataset using a span corruption objective. In this paper, several model variants were tested: T5-small (60 million parameters), T5-base (220 million parameters), and T5-large (770 million parameters).

In terms of VLM applications, studies have been conducted on the use of VLMs in biomedical image analysis by integrating LLMs with Computer-Aided Diagnosis (CAD) networks for clinical applications. For instance, Wang et al. combined existing LLMs with CAD networks, demonstrating how these models can be applied to diagnostic tasks \cite{wangChatCADInteractiveComputerAided2023}. Similarly, Yan et al. explored the performance of ChatGPT-4V on simple medical Visual Question Answering (VQA) tasks. Although the results are promising, they find it unsuitable for real-world diagnostic scenarios due to its inability to efficiently handle complex medical visual tasks ~\cite{yanMultimodalChatGPTMedical2023}.
\subsection{Challenges of AI Models in Neuroimaging}
To date, the application of these models in the field of neuroimaging remains underexplored. Neuroimaging data are characterized by high complexity and variability, making it challenging to directly transfer the methodologies developed for other biomedical fields to this domain. Although models like LLaVA-Med, Biomed-GPT, and Geneformer have been successfully applied in fields such as radiology, pathology, and general medical imaging analysis, their use in specialized areas like Alzheimer's disease diagnosis has not yet been thoroughly investigated. For instance, Med-PaLM 2, one of the leading generalist AI models, has achieved excellent results in broader biomedical tasks such as MedQA, but its application in specific domains like Alzheimer's disease diagnosis remains unexplored. Therefore, the performance of these models in such diagnostic contexts could be poor\cite{vanLargeVisualLanguage2024,tuGeneralistBiomedicalAI2023,wangChatCADInteractiveComputerAided2023}.
\subsection{Relevant Datasets}
The most commonly used datasets for training Vision-Language Models (VLMs) in automatic diagnosis and diagnostic report generation are MIMIC-CXR, IU-Xray, and CXR-PRO. These datasets primarily focus on chest radiographic images and include comprehensive textual diagnostic reports \cite{hartsockVisionLanguageModelsMedical2024}. In the field of Alzheimer’s disease, notable datasets include the Alzheimer’s Disease Neuroimaging Initiative (ADNI) \cite{petersenAlzheimerDiseaseNeuroimaging2010}, Open Access Series of Imaging Studies (OASIS) \cite{marcusOpenAccessSeries2010}, and Australian Imaging, Biomarkers and Lifestyle (AIBL) \cite{AIBLStudy2024}. These datasets contain a significant amount of multimodal data, such as magnetic resonance imaging, MRI, and positron emission tomography (PET) images, alongside structured data related to cognitive tests, disease status evaluations, genetic biomarkers and volumetric segmentations extracted using FreeSurfer \cite{fischlFreeSurfer2012}. However, unlike radiology datasets, such as MIMIC-CXR and IU-Xray, these comprehensive neuroimaging datasets do not include complete diagnostic reports associated with the medical images, an essential element for fine-tuning VLMs and subsequent validation. This is precisely one of the main challenges we aim to address in this work.

In summary, the main differences between our paper and the related work lie in three aspects. First, we propose the generation of synthetic diagnostic reports to address the lack of textual descriptions in neuroimaging datasets, particularly for the training of advanced language models. Second, our approach emphasizes the integration of visual transformer and language transformer models, which enhances the diagnostic capabilities through a more effective multimodal data fusion. Finally, we introduce a methodology for selecting relevant structured data from datasets such as OASIS-4, enabling the generation of clinically relevant and accurate synthetic reports.

\section{OASIS Datasets}
\label{sec:datasets}
The Open Access Series of Imaging Studies (OASIS) dataset was selected for its accessibility and the completeness of the data it contains. Developed through research initiatives conducted at the Washington University Knight Alzheimer Disease Research Center, this dataset includes comprehensive clinical and neuroimaging information and comprises four cohorts: OASIS-1 and OASIS-2, which mainly contain MR images and basic demographic information without detailed structured data. OASIS-3 and OASIS-4, instead, also include clinical and cognitive assessments, as well as FreeSurfer segmentations, in addition to MR images\cite{marcusOpenAccessSeries2007}.

\subsection{OASIS-1}
OASIS-1 \cite{marcusOpenAccessSeries2007} is a cross-sectional dataset comprising 416 subjects, aged between 18 and 96 years. Each subject underwent 3 to 4 T1-weighted MRI scans, acquired in a single scanning session. Among these subjects, 100 individuals aged over 60 have been diagnosed with very mild to moderate Alzheimer's disease. A subgroup of 20 patients underwent a second imaging session within 90 days of the initial visit. The dataset includes 434 MRI T1w sessions along with basic clinical and demographic data such as age, gender, and cognitive status. The images were preprocessed and segmented using the FreeSurfer software to extract brain volumes.

\subsection{OASIS-2} OASIS-2 \cite{marcusOpenAccessSeries2010a} is a longitudinal dataset comprising 150 subjects, aged between 60 and 96 years. Each subject underwent multiple scans in two or more visits, at least one year apart, for a total of 373 imaging sessions. Among these subjects, 72 remained nondemented throughout the study, 64 were diagnosed as demented at the initial visit (including 51 with mild to moderate Alzheimer’s disease), and 14 transitioned from a nondemented state to a diagnosis of dementia in subsequent visits. The dataset includes detailed clinical and cognitive data, such as neuropsychometric and diagnostic evaluations like Clinical Dementia Rating (CDR) and Mini-Mental State Examination (MMSE), enabling the monitoring of disease progression over time.

\subsection{OASIS-3}
The OASIS-3 \cite{lamontagneOASIS3LongitudinalNeuroimaging2019} dataset is a retrospective collection of data from 1379 participants gathered over the course of 30 years. The cohort includes 755 cognitively normal adults and 622 individuals with various degrees of cognitive decline, aged 42 to 95 years.

\begin{itemize} 
    \item \textbf{Imaging}: OASIS-3 contains 2842 MRI sessions with various sequences, such as T1w, T2w, FLAIR, ASL, SWI, DTI, and resting-state BOLD. Additionally, there are 2157 PET sessions, using tracers like PIB, AV45, and FDG to study brain metabolism and amyloid accumulation;
    \item \textbf{Clinical and cognitive data:} The clinical evaluations include standardized measures such as the CDR and cognitive tests like the MMSE. Genetic data, such as APOE status, are also included;
    \item \textbf{Processing}: MR imaging data were processed using the FreeSurfer software to obtain volumetric segmentations, including the hippocampus, amygdala, and other brain regions involved in neurodegenerative diseases \cite{laaksoVolumesHippocampusAmygdala1995}. 
\end{itemize}
\subsection{OASIS-4}
The OASIS-4 \cite{Koenig2020} dataset includes 663 participants aged 21 to 94 years, evaluated for memory disorders or dementia. OASIS-4 represents an independent dataset focused on clinical, cognitive, and neuroimaging data.

\begin{itemize} 
    \item \textbf{Imaging:} OASIS-4 includes 676 MRI sessions for structural brain analysis, but unlike OASIS-3, it does not include PET sessions;
    \item \textbf{Clinical and cognitive data: }Clinical evaluations include neuropsychometric tests and biomarker measurements. Diagnostic tools such as the CDR and MMSE are also used in this cohort;
    \item \textbf{Processing:} The images of the patients in this dataset have been processed using FreeSurfer to extract volumetric segmentations of areas involved in neurodegenerative diseases.
    \end{itemize}
The most informative datasets for model training are OASIS-3 and OASIS-4. In the development of this work, particular reference will be made to the OASIS-4 dataset, as OASIS-3 is more suitable for longitudinal studies. The Table \ref{tab:oasis_cohorts} summarizes the content of the two datasets.

\begin{table*}[ht]
  \caption{Characteristics of OASIS-3 and OASIS-4 cohorts.}
  \label{tab:oasis_cohorts}
  \centering
  \begin{tabularx}{\linewidth}{@{\extracolsep{\fill}}c c c}
    \toprule
    \textbf{Characteristic} & \textbf{OASIS-3} & \textbf{OASIS-4} \\
    \midrule
    Number of participants & 1379 & 663 \\
    Participant age range & 42-95 years & 21-94 years \\
    Data types & Longitudinal, neuroimaging, clinical, cognitive & Clinical, neuropsychometric, MRI \\
    MRI sessions & 2842 & 676 \\
    PET sessions & 2157 & None \\
    Focus & Aging and Alzheimer's disease & Memory disorders, dementia \\
    \bottomrule
  \end{tabularx}
\end{table*}

\section{Methodology}
\label{sec:methods}
\begin{figure}[ht]
    \centering
    \includegraphics[width=\linewidth]{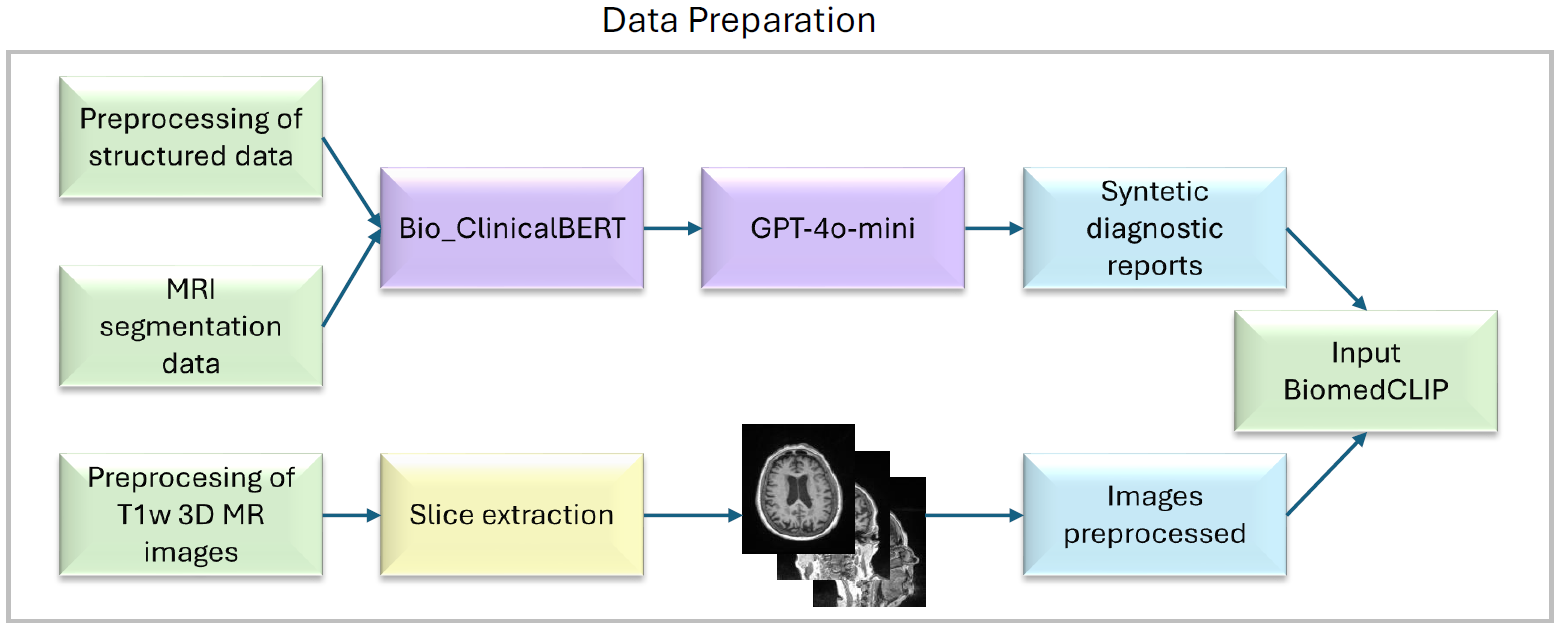}  % Usa \linewidth per adattare la figura alla larghezza della colonna
    \caption{Data preparation workflow for structured data, MR images, and segmentation volumes.}
    \label{fig:data_prepraration}
\end{figure}

\begin{figure*}[t] 
    \centering
    \includegraphics[width=\textwidth]{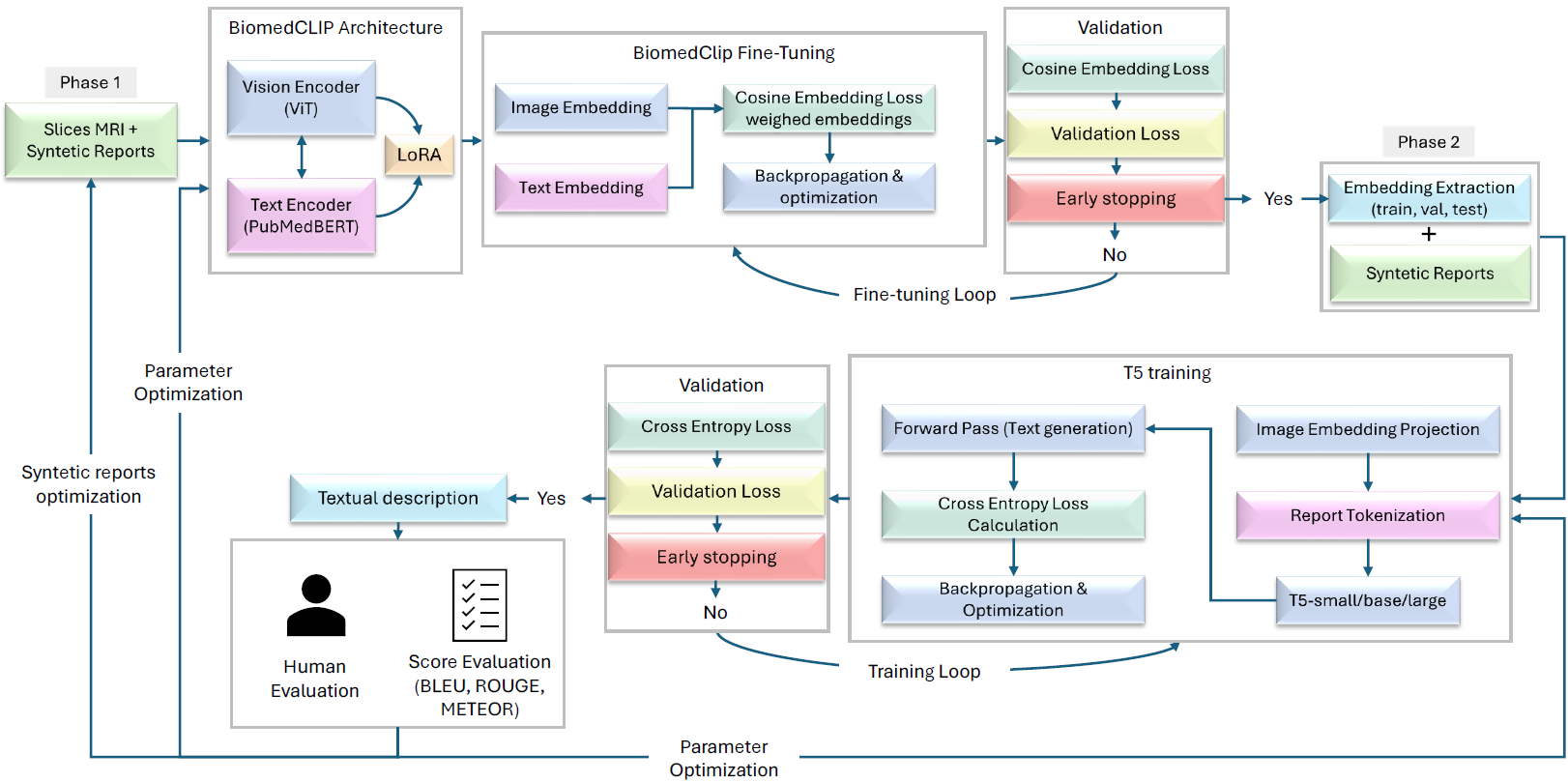}  % Usa \textwidth per adattare la figura alla larghezza totale delle due colonne
    \caption{Overview of the methodology, showing data preprocessing, model fine-tuning, and evaluation phases.}
    \label{fig:methods_overview_flowchart}
\end{figure*}

This section illustrates the methodology adopted to address the research questions. The workflow is organized into three phases: data preprocessing, model training, and validation. The process starts with the preprocessing of structured clinical data and neuroimaging data, followed by the generation of synthetic reports from these data, which will be used as ground truth for model training and validation. Subsequently, state-of-the-art VLMs and LLMs are integrated for fine-tuning. Finally, the generated reports are validated using natural language generation metrics to ensure clinical relevance and accuracy. Fig. \ref{fig:data_prepraration} shows the data preparation process that will subsequently be used by the state-of-the-art VLM BiomedCLIP and the LLM T5, while Fig. \ref{fig:methods_overview_flowchart} provides a complete overview of the proposed methodology for model training, validation, and testing.

\subsection{Preprocessing of Structured Data}
The structured data from the OASIS-4 dataset required extensive preprocessing to address missing values and normalize variables. In some cases, patients had incomplete entries in the structured data, while in others, corresponding FreeSurfer volumetric measurements were missing. To ensure data consistency and reliability, only patients with complete clinical and imaging data, obtained within one year of the MRI visit, were selected, resulting in a final cohort of 468 patients. The clinical, neuropsychometric, and imaging datasets were then merged using the patient ID as a key, with values aligned according to visit days. Subsequently, files containing clinical, neuropsychometric, and cognitive scores were used to create a cohesive dataset for the subsequent generation of synthetic diagnostic reports. The structured data selected for preliminary tests includes values such as CDR, Sumbox (sum of boxes), MMSE, and final diagnosis, providing detailed clinical information on the cognitive status of patients. The created dataset also includes volumetric measurements of brain areas such as the left hippocampus and right hippocampus. The combination of cognitive, clinical, and anatomical data offers a comprehensive overview of each patient’s health status, making it possible to generate accurate synthetic reports.

\subsection{Generation of Synthetic Diagnostic Reports} Since the available neuroimaging datasets do not provide complete textual reports associated with MR images, it was necessary to generate synthetic reports to serve as ground truth for the training and validation of the models. To this end, the GPT-4o-mini API was used in combination with the Bio\_ClinicalBERT model to generate clinically relevant texts. Bio\_ClinicalBERT is a variant of the BERT (Bidirectional Encoder Representations from Transformers) model, trained on corpora of clinical texts to improve its effectiveness in understanding medical domain concepts. In combination with GPT-4o-mini, which facilitates text generation, Bio\_ClinicalBERT enriched the descriptions with relevant clinical details based on the available structured data \cite{alsentzerPubliclyAvailableClinical2019,leeBioBERTPretrainedBiomedical2020}. The choice of GPT-4o-mini as the report generation model was due to a compromise between the cost of the API usage and the quality of the synthetic reports generated.

The generation process involves several key steps. First, the structured data underwent a phase of normalization and feature extraction. Specifically, predefined ranges were set to categorize clinical variables such as MMSE, CDR, and hippocampal volumes into qualitative descriptions to ensure consistency in report generation. Following this preprocessing phase, the structured data were tokenized using the Bio\_ClinicalBERT tokenizer. These tokenized inputs were then passed through Bio\_ClinicalBERT to extract relevant features. A specific prompt was created to guide GPT-4o-mini in generating synthetic texts. The prompt instructed the model to generate reports of 100-150 words, emphasizing appropriate medical terminology, with a temperature setting of 0.7 to better mimic real clinical scenarios.
We note that while clinical validation of the synthetic reports by expert clinicians is crucial for improving the model’s performance, it is not the key focus of this paper but will be part of our future work.

\subsection{Preprocessing of T1-Weighted MR Images} The T1-weighted MR images were preprocessed to provide visual input to the BiomedCLIP model. Preprocessing was applied to the entire 3D volume of each MRI scan. The images were initially converted to the RAS (Right, Anterior, Superior) orientation for standardization. Subsequently, the intensity values were normalized to fall within the range of 0 to 1 using the 2nd and 98th percentiles to clip intensity values and reduce outliers \cite{parkWhiteMatterHyperintensities2021}. A Gaussian filter with a value of 0.5 was used to reduce noise in the images. This value was chosen to balance noise reduction with the preservation of relevant anatomical details. Higher values compromise the definition of brain structures, while lower values are less effective in noise reduction. Additionally, a gamma correction with an exponent of 0.8 was applied to enhance the contrast of the images. Finally, background noise was removed by setting intensity values below the 1st percentile threshold to zero, retaining only the diagnostically significant regions of the image. After preprocessing, 2D slices were extracted from the 3D volumes along the axial, coronal, and sagittal planes, selecting a fixed number of central slices for each orientation to ensure comprehensive brain representation.

\subsection{Training and Validation of the Models}
The training process for automatic text report generation consists of two main phases executed on workstation equipped with an NVIDIA GeForce RTX 4090 GPU:
\begin{itemize}
    \item \textbf{Fine-tuning of the BiomedCLIP model} to generate embeddings from the MRI slices of the patients;
    \item \textbf{Fine-tuning of the T5 model} to generate textual diagnostic reports based on the numerical embeddings previously extracted.
\end{itemize}

\subsubsection{Fine-Tuning and Validation of the BiomedCLIP Multimodal Model}
The model was adapted to the OASIS dataset and synthetic reports through the PEFT (Parameter-Efficient Fine-Tuning) library \cite{xuParameterEfficientFineTuningMethods2023}. Specifically, the LoRA (Low-Rank Adaptation)  \cite{huLoRALowRankAdaptation2021} technique was used, which allows for efficient model fine-tuning by modifying only a small portion of the parameters. Specifically, LoRA was applied to the model’s \texttt{q\_proj} and \texttt{v\_proj} modules. The LoRA configurations used for fine-tuning include an adaptation factor \( r = 10 \), a parameter \( \textit{lora\_alpha} = 32 \), and a dropout value of \( \textit{lora\_dropout} = 0.3 \).

For training, three distinct datasets were created using scikit-learn's \textit{train\_test\_split} function: a training set (70\%), a validation set (20\%), and a test set (10\%), generated from MR images and synthetic reports of the OASIS dataset. The multiple MRI slices of each subject are associated with the same synthetic report, allowing the model to learn to associate various brain regions with the overall medical description. The multimodal inputs (images and text) are preprocessed using the BiomedCLIP processor, necessary for tokenizing the reports and converting the images into tensors.
During the training phase, which was conducted for 100 epochs with a batch size of 64, the embeddings generated for images and text were compared using a \textit{CosineEmbeddingLoss} function. This loss function is based on cosine similarity and helps the model learn to align the image and text embeddings. The similarity target value was set to 1 to maximize the similarity between the image and text embeddings. An attention-weighted mechanism is also used to compute the weighted average of embeddings for each patient, aggregating information obtained from multiple MRI slices. This process enables the model to learn multimodal relationships between brain images and textual reports, giving more importance to slices that are more relevant to the report content. Model optimization is carried out using the \textit{AdamW} algorithm, with an initial learning rate of \(1e - 5\) and a weight decay of 0.01. Early stopping with a patience of 10 epochs and gradient clipping with a maximum norm of 1.0 were also implemented to prevent overfitting and stabilize the training. Validation is performed using the same loss function, and attention-weighted mechanism, calculating the similarity score between the embeddings for each patient. The learning rate is dynamically adjusted using the \textit{ReduceLROnPlateau} scheduler, which reduces the learning rate when the validation loss reaches the plateau. Finally, the aggregated embeddings of images and text are extracted and saved for each patient to be subsequently used by the T5 model.

\subsubsection{Fine-Tuning and Validation of the T5 Model}
Below is the procedure for using the large variant of the T5 model. The methodology was also tested with the small and base variants by selecting the corresponding model size during the configuration phase. The fine-tuning of the T5 model was performed using the multimodal embeddings generated by BiomedCLIP and diagnostic reports. The final objective is to generate complete clinical reports based on MRI representations and combined textual information. The optimization process of T5 consists of several phases. To improve the model's robustness and generalization, data augmentation techniques were applied during fine-tuning. Data augmentation is applied to both image embeddings and text reports, as follows:
\begin{itemize}
    \item \textbf{Embedding augmentation}: random Gaussian noise (with a standard deviation of 0.1) is added to the image embeddings to handle small fluctuations in the input data;
    \item \textbf{Text augmentation:} 10\% of the words in the reports are replaced with synonyms from the \textit{NLTK} \textit{WordNet} database \cite{NaturalLanguageProcessing,princeton2010wordnet}, selected using a random sampling strategy.
\end{itemize}
Initially, the embeddings extracted from BiomedCLIP are combined with the synthetic textual reports of each patient. Before integration into T5, a projection network was introduced to adapt the dimensionality of the embeddings produced by BiomedCLIP (512 dimensions) to the dimensions required by T5 large (1024 dimensions). The projection layer consists of a linear layer, followed by a LayerNorm layer and a dropout layer (rate = 0.2) to prevent overfitting and normalize the projected inputs. The diagnostic reports are tokenized and encoded with the T5 tokenizer, maintaining a maximum length of 512 tokens.

During fine-tuning the training data is organized in batches of 4 samples. To prevent overfitting, early stopping criteria is applied, terminating the training if no improvement in the validation loss is observed for 15 consecutive epochs. The \textit{CrossEntropyLoss} function is used to compare the reports generated by the model with the reference reports. For model optimization, the \textit{AdamW} algorithm is used with a learning rate of \(1e-4\) and a weight decay of 0.05. The \textit{ReduceLROnPlateau} scheduler is used to dynamically reduce the learning rate based on validation loss values with a patience of 10 epochs and a factor of 0.5. Additionally, gradient clipping with a maximum norm of 1.0 is applied to stabilize training. The report generation phase uses sampling techniques such as \(top-k (k=10)\) and \(top-p (p=0.9)\) with a temperature of 0.6 to balance diversity and coherence in the generated reports.
\subsection{Testing and Evaluation of the T5 Model}
The model performance evaluation was carried out using natural language generation metrics. Specifically, Bilingual Evaluation Understudy (BLEU), Metric for Evaluation of Translation with Explicit Ordering (METEOR), and Recall-Oriented Understudy for Gisting Evaluation (ROUGE) were selected \cite{hartsockVisionLanguageModelsMedical2024,jiVisionLanguageModelGenerating2024}.
\subsubsection{BLEU Score}

This metric was originally used to evaluate the quality of machine-generated translations by comparing them to one or more reference translations. The BLEU score calculates a precision-based metric by counting the number of n-grams (continuous sequences of n items) in the generated text that match any reference translation. In the evaluation of the reports, precision was calculated for n-grams with values ranging from 1 to 4 (BLEU-1, BLEU-2, BLEU-3, BLEU-4):
\begin{equation}
    \text{Precision}(n) = \frac{\text{\# overlapping n-grams}}{\text{\# all n-grams (model-generated)}}
    \label{eq:precision}
\end{equation}
The BLEU-n score formula is reported below:
\begin{equation}
    \text{BLEU-}n = BP \times \frac{1}{n} \exp\left(  \sum_{k=1}^{n} \log \left( \text{Precision}(k) \right) \right)
\end{equation} 
Where BP refers to the brevity penalty and is calculated as:   
\begin{equation}
BP = 
\begin{cases}
1 & \text{if } c \geq r \\
e^{(1 - \frac{r}{c})} & \text{if } c < r
\end{cases}
\end{equation}
Where \( c \) is the length of the model-generated text, and \( r \) is the length of the reference text. The BLEU score ranges from 0 to 1, with values closer to 1 indicating better agreement with the reference text.
In the code implementation, the BLEU score was calculated using the \textit{NLTK} library, specifically by using the \textit{sentence\_bleu} function \cite{NaturalLanguageProcessing}. The BLEU score for 1 to 4 n-grams was computed both for each individual patient with respect to the reference text and at the corpus level.
\subsubsection{ROUGE Score}
ROUGE is a set of metrics used to evaluate the overlap between the model-generated text and the human reference text, where ROUGE-n measures the overlap of n-grams between the two. ROUGE captures both precision and recall, providing a more balanced evaluation. The metric is calculated as follows:
\begin{equation}
\text{ROUGE-}n = \frac{\text{\# overlapping n-grams}}{\text{\# all n-grams in the reference text}}
\end{equation}
ROUGE-L, on the other hand, focuses on measuring the longest common subsequence between the model-generated text \( Y \) and the reference text \( X \). It is calculated using the following formula:
\begin{equation}
    \text{ROUGE-L} = \frac{(1 + \beta^2) \times R \times P}{(R + P \times \beta^2)}
\end{equation}
Where \( R = \frac{\text{LCS}(X, Y)}{m} \) and \( P = \frac{\text{LCS}(X, Y)}{n} \). Here, \( m \) is the length of \( X \), \( n \) is the length of \( Y \), and \(\text{LCS}(X, Y)\) is the length of the longest common subsequence between \( X \) and \( Y \). The parameter \(\beta\) controls the weight given to precision (\( P \)) and recall (\( R \)) based on the specific task and their relative importance.
The ROUGE scores range from 0 to 1, where 1 indicates a perfect similarity between the generated text and the reference text.
In the code implementation, the \textit{RougeScorer} function from the \textit{rouge\_score} library was used. Additionally, the ROUGE score was computed at the corpus level using \textit{BootstrapAggregator} \cite{lin2004rouge}.
\subsubsection{METEOR Score}
This evaluation metric is based on the harmonic mean of unigram precision and recall, with recall weighted higher than precision. This metric is designed to be less strict compared to other metrics and considers the fluency and meaning of generated text. Specifically, it is calculated as follows:
\begin{equation}
    \text{METEOR} = \frac{10 \times P \times R}{R + 9 \times P} \times (1 - \text{Penalty})
\end{equation}
\noindent where
\begin{equation}
    R = \frac{\text{\# overlapping 1-grams}}{\text{\# 1-grams in a reference text}}
\end{equation}
\begin{equation}
       P = \frac{\text{\# overlapping 1-grams}}{\text{\# 1-grams in a model-generated text}}
\end{equation}
\begin{equation}
    \text{Penalty} = \frac{1}{2} \times \left( \frac{\text{\# chunks}}{\text{\# overlapping 1-grams}} \right)^{3}
\end{equation}
and chunks are groups of adjacent 1-grams in the model-generated text that overlap with adjacent 1-grams in the reference text. The METEOR score ranges from 0 to 1 and was computed using the NLTK library. In this study, the score was computed at the level of individual reports, and then the average was taken. The METEOR score is specifically designed to provide a detailed evaluation at the sentence level, unlike other metrics such as BLEU and ROUGE. This characteristic stems from the fact that METEOR not only considers word matches but also incorporates advanced linguistic features, such as synonym handling, stemming, and semantic matching \cite{banarjee2005}.

\section{Results and Discussion}
\label{sec:results}
This section presents the preliminary results obtained using the three versions of the T5 model: small, base, and large, evaluated on the OASIS-4 dataset as described in Section \ref{sec:datasets}. Specifically, the corpus-level scores (Table \ref{tab:corpus_scores}) for the following metrics are reported: BLEU-1, BLEU-2, BLEU-3, BLEU-4, ROUGE-1, ROUGE-2, ROUGE-L (using the F1 score), and METEOR.
\begin{table*}[hbt!]
  \caption{Corpus Scores for T5 Model Variants}
  \label{tab:corpus_scores}
  \centering
  \begin{tabularx}{\linewidth}{@{\extracolsep{\fill}}l c c c c c c c c}
    \toprule
    \textbf{Model} & \textbf{BLEU-1} & \textbf{BLEU-2} & \textbf{BLEU-3} & \textbf{BLEU-4} & \textbf{ROUGE-1} & \textbf{ROUGE-2} & \textbf{ROUGE-L} & \textbf{METEOR} \\
    \midrule
    T5-Small & 0.5617 & 0.3758 & 0.2632 & 0.1858 & 0.5864 & 0.2542 & 0.3625 & 0.4167 \\
    T5-Base  & 0.5512 & 0.3699 & 0.2604 & 0.1820 & 0.5784 & 0.2517 & 0.3660 & 0.4271 \\
    T5-Large & 0.5519 & 0.3711 & 0.2604 & 0.1827 & 0.5908 & 0.2565 & 0.3719 & 0.4163 \\
    \bottomrule
  \end{tabularx}
\end{table*}

%\begin{table*}[htbp!]
 % \caption{Average Scores for T5 Model Variants}
  %\label{tab:avg_scores}
  %\centering
  %\begin{tabular}{l@{\hskip 0.265cm}c@{\hskip 0.265cm}c@{\hskip 0.265cm}c@{\hskip 0.265cm}c@{\hskip 0.265cm}c@{\hskip 0.265cm}c@{\hskip 0.265cm}c@{\hskip 0.265cm}c}
   % \toprule
   % \textbf{Model} & \textbf{BLEU-1\textsuperscript{a}} & \textbf{BLEU-2\textsuperscript{a}} & \textbf{BLEU-3\textsuperscript{a}} & \textbf{BLEU-4\textsuperscript{a}} & \textbf{ROUGE-1\textsuperscript{a}} & \textbf{ROUGE-2\textsuperscript{a}} & \textbf{ROUGE-L\textsuperscript{a}} & \textbf{METEOR\textsuperscript{a}} \\
   % \midrule
   % T5-Small & 0.5516 & 0.3671 & 0.2543 & 0.1759 & 0.5867 & 0.2544 & 0.3633 & 0.4059 \\
   % T5-Base  & 0.5396 & 0.3595 & 0.2498 & 0.1701 & 0.5788 & 0.2514 & 0.3666 & 0.3962 \\
   % T5-Large & 0.5414 & 0.3624 & 0.2515 & 0.1725 & 0.5908 & 0.2561 & 0.3717 & 0.4058 \\
   %%\end{tabular}
%\end{table*}
The results show similar performance across the T5 model variants, with some slight differences. The T5-small model achieved higher BLEU scores compared to T5-base and T5-large, particularly for BLEU-1 (0.5617) and BLEU-4 (0.1858). This suggests that the smaller model might be more effective at generating n-grams that match the reference text. On the other hand, the T5-large model showed superior performance in terms of ROUGE-1 (0.5908), ROUGE-2 (0.2565), and ROUGE-L (0.3719). These results indicate that the larger model might produce outputs that better capture the overall structure and coherence of the text, generating longer sequences that align more closely with the reference text. As for the METEOR score, the T5-small and T5-large models achieved similar results (0.4167 and 0.4163, respectively), while at the corpus level, the T5-base model outperformed the other two (0.4271).

\begin{figure*}[t] 
    \centering
    \includegraphics[width=\textwidth]{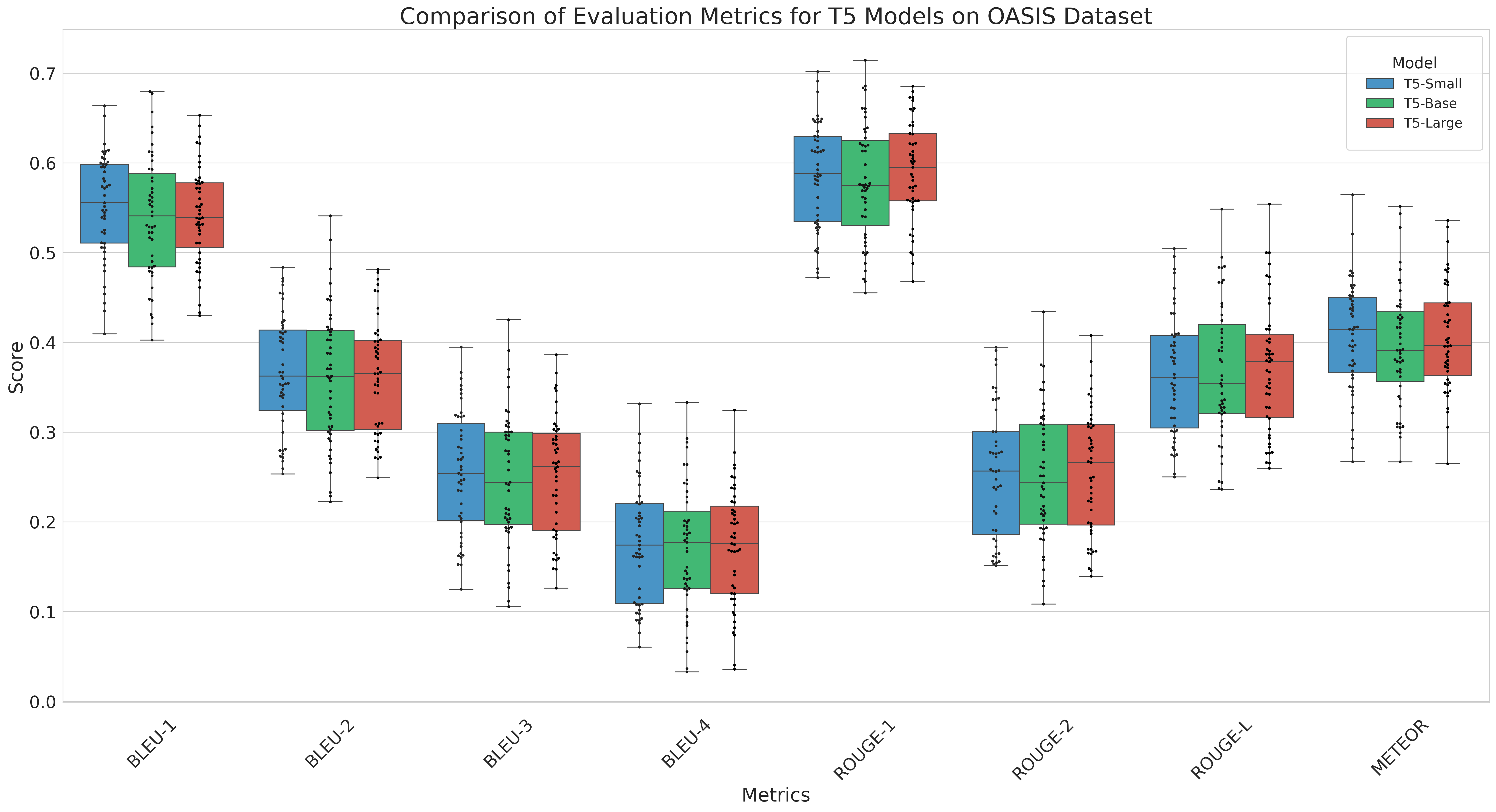}  
    \caption{Boxplot comparison of BLEU, ROUGE, and METEOR evaluation metrics for T5-Small, T5-Base, and T5-Large models on the OASIS-4 dataset.}
    \label{fig:boxplots}
\end{figure*}

The boxplot in Fig. \ref{fig:boxplots} illustrates the quartile distribution of the evaluation metrics, and Table \ref{tab:quartiles} provides a detailed analysis of the main metrics. For the BLEU-4 score, although the median values are comparable across all three models, the 75th percentile is slightly higher for T5-Small (0.2207) compared to the other models. However, as highlighted in the boxplot \ref{fig:boxplots}, T5-Small shows greater variability than T5-Base and T5-Large, indicating less stable performance. Regarding ROUGE-1 and ROUGE-L, the median and third quartile values for T5-Large (0.5951 and 0.6324 for ROUGE-1, 0.3784 and 0.4092 for ROUGE-L) demonstrate the model’s ability to generate text that better aligns with the structure and coherence of the reference text. These results are further confirmed by Fig. \ref{fig:boxplots}, where T5-Large shows narrower interquartile ranges and fewer extreme values, leading to more consistent and robust performance compared to T5-Small and T5-Base.
\begin{table}[htbp]
  \caption{Distribution of Evaluation Metrics Across T5 Model Variants (Quartiles)}
  \label{tab:quartiles}
  \centering
  \begin{tabularx}{\linewidth}{@{\extracolsep{\fill}}l c c c c}
    \toprule
    \textbf{Metric} & \textbf{Model} & \textbf{25\%} & \textbf{50\% (Median)} & \textbf{75\%} \\
    \midrule
    BLEU-1 & T5-Small & 0.5106 & 0.5556 & 0.5982 \\
           & T5-Base  & 0.4842 & 0.5408 & 0.5881 \\
           & T5-Large & 0.5054 & 0.5388 & 0.5777 \\
    \midrule
    BLEU-4 & T5-Small & 0.1093 & 0.1741 & 0.2207 \\
           & T5-Base  & 0.1259 & 0.1773 & 0.2119 \\
           & T5-Large & 0.1202 & 0.1758 & 0.2175 \\
    \midrule
    ROUGE-1 & T5-Small & 0.5346 & 0.5878 & 0.6297 \\
            & T5-Base  & 0.5300 & 0.5752 & 0.6247 \\
            & T5-Large & 0.5576 & 0.5951 & 0.6324 \\
    \midrule
    ROUGE-L & T5-Small & 0.3046 & 0.3604 & 0.4074 \\
            & T5-Base  & 0.3207 & 0.3541 & 0.4196 \\
            & T5-Large & 0.3163 & 0.3784 & 0.4092 \\
    \midrule
    METEOR  & T5-Small & 0.3660 & 0.4142 & 0.4501 \\
            & T5-Base  & 0.3566 & 0.3913 & 0.4349 \\
            & T5-Large & 0.3632 & 0.3962 & 0.4440 \\
    \bottomrule
  \end{tabularx}
\end{table}

The Table \ref{tab:ratio_comparison} shows a comparison of the performance of the T5 model in its three variants (small, base, and large) based on different dataset splitting ratios (Train/Validation/Test).
The results indicate that the 70/20/10 split provides the best overall results for the BLEU-4, ROUGE-1, ROUGE-2, ROUGE-L, and METEOR metrics, followed by the 60/30/10 configuration and finally the 80/10/10 configuration, which demonstrates the worst performance among the three ratios. The 70/20/10 split configuration provides overall better results as shown by the following metrics: BLEU-4 (T5-Large: 0.1827), ROUGE-1 (T5-Large: 0.5908), ROUGE-2 (T5-Large: 0.2565), ROUGE-L (T5-Large: 0.3719), and METEOR (T5-Base: 0.4271).
This indicates that a split with a higher percentage of training data and a balanced proportion of validation data leads to more promising results. In contrast, a smaller amount of validation data, as seen in the 80/10/10 configuration, hinders overfitting control, reducing the model’s generalization capabilities.
In conclusion, the table highlights that the 70/20/10 configuration is optimal for all T5 model variants, as it ensures a better balance between training and validation data, resulting in improved report accuracy and consistency.

\begin{table*}[htb]
  \caption{Comparison of Metrics for Different Train/Val/Test Ratios}
  \label{tab:ratio_comparison}
  \centering
  \begin{tabularx}{\linewidth}{@{\extracolsep{\fill}}l c c c c c c c c c}
    \toprule
    \textbf{Model} & \textbf{Ratio} & \textbf{BLEU-1} & \textbf{BLEU-2} & \textbf{BLEU-3} & \textbf{BLEU-4} & \textbf{ROUGE-1} & \textbf{ROUGE-2} & \textbf{ROUGE-L} & \textbf{METEOR} \\
    \midrule
    T5-Small & 60/30/10 & 0.5375 & 0.3522 & 0.2396 & 0.1621 & 0.5624 & 0.2329 & 0.3477 & 0.3837 \\
    T5-Small & 70/20/10 & 0.5617 & 0.3758 & 0.2632 & 0.1858 & 0.5864 & 0.2542 & 0.3625 & 0.4167 \\
    T5-Small & 80/10/10 & 0.5384 & 0.3518 & 0.2409 & 0.1660 & 0.5602 & 0.2318 & 0.3433 & 0.3874 \\
    \midrule
    T5-Base & 60/30/10 & 0.5588 & 0.3663 & 0.2498 & 0.1707 & 0.5828 & 0.2441 & 0.3602 & 0.3980 \\
    T5-Base & 70/20/10 & 0.5512 & 0.3699 & 0.2604 & 0.1820 & 0.5784 & 0.2517 & 0.3660 & 0.4271 \\
    T5-Base & 80/10/10 & 0.5572 & 0.3647 & 0.2509 & 0.1723 & 0.5732 & 0.2361 & 0.3541 & 0.3943 \\
    \midrule
    T5-Large & 60/30/10 & 0.5461 & 0.3535 & 0.2379 & 0.1586 & 0.5690 & 0.2338 & 0.3445 & 0.3867 \\
    T5-Large & 70/20/10 & 0.5519 & 0.3711 & 0.2604 & 0.1827 & 0.5908 & 0.2565 & 0.3719 & 0.4163 \\
    T5-Large & 80/10/10 & 0.5552 & 0.3624 & 0.2432 & 0.1624 & 0.5701 & 0.2314 & 0.3447 & 0.3922 \\
    \bottomrule
  \end{tabularx}
\end{table*}

\begin{figure*}[htb!]
  \centering
  \includegraphics[width=\textwidth]{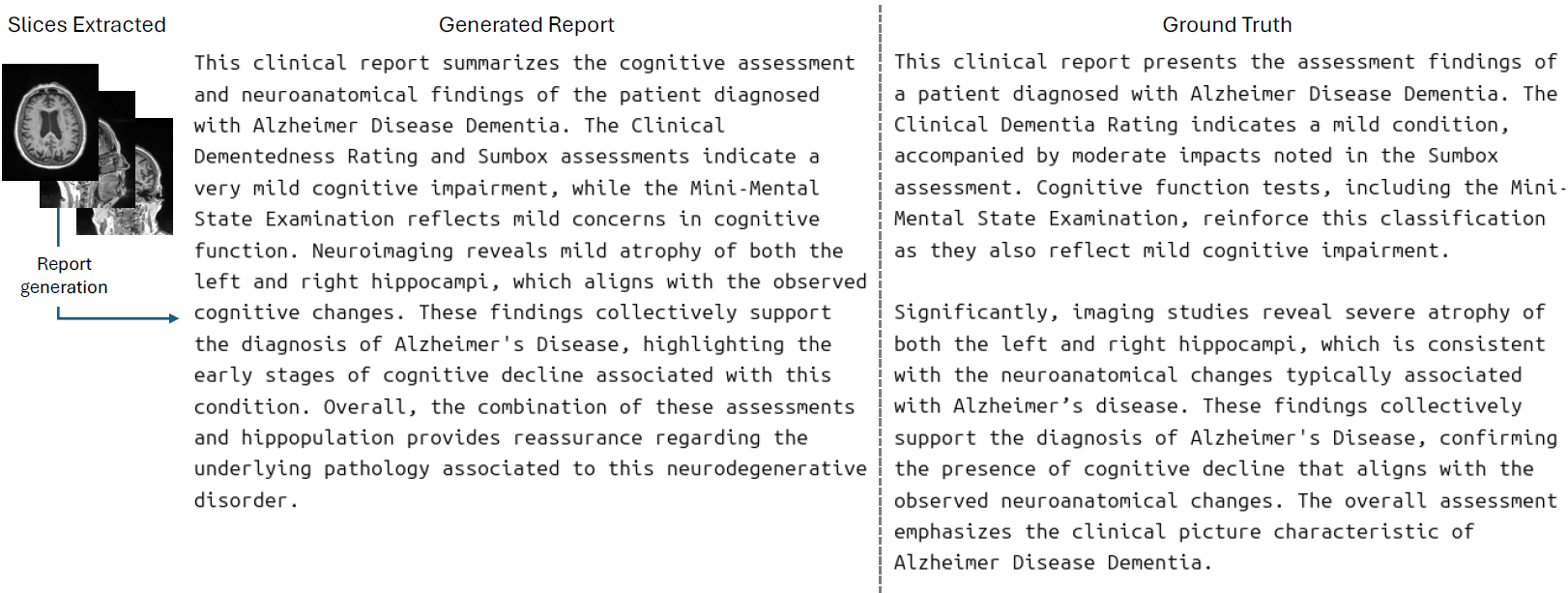}
  \caption{Generation of a textual report using BiomedCLIP and T5-large and comparison with the ground truth.}
  \label{fig:report_generation}
\end{figure*}

Fig. \ref{fig:report_generation} shows an example of a text report generated using the T5-large model, which achieved the best results among its outputs in terms of ROUGE-1 (0.6854), ROUGE-2 (0.4076), ROUGE-L (0.5540), and BLEU-4 (0.3245) scores. Comparing the generated report with the reference text, several observations can be made. The overall structure of the generated report closely aligns with the ground truth, adhering to a format similar to that of the synthetic reports. In terms of diagnostic accuracy, the generated report correctly identifies the diagnosis of Alzheimer's Disease Dementia, as reported in the reference text. However, there are some differences. For instance, the model generated the phrase ``very mild cognitive impairment", whereas the ground truth indicates the condition as ``mild". Additionally, the Sumbox assessment is described as ``very mild" in the generated text, while in the ground truth it is reported as having a ``moderate impact". There are also differences in the description of hippocampal atrophy. In fact, the generated report mentions ``mild atrophy", while the reference text indicates ``severe atrophy". A lexical inaccuracy was also observed, as the model generated the non-existent word ``hippopulation",  which likely resulted from an unintended fusion of two words.

In summary, the generated report shows an overall structure that is consistent with the ground truth; however, discrepancies and clinical details highlight the need for further refinement to improve its overall accuracy.

\section{Conclusion and Future Work}
\label{sec:conclusion}
This study proposes an innovative approach to address the limitations of VLMs and LLMs in the field of neuroimaging diagnostics by using structured clinical data and MR images to generate diagnostic reports. BiomedCLIP and T5 were utilized with OASIS-4 data, demonstrating the potential of creating a framework capable of generating diagnostic reports for Alzheimer’s disease. Our preliminary results indicate that the proposed methodology can effectively integrate visual information and textual data to produce clinically detailed descriptions for neurodegenerative diseases such as Alzheimer’s.

However, there are some limitations that should be noted for an accurate interpretation of the results and to guide future research. One key limitation is the sample size used in this study. The study only employed data from the OASIS-4 cohort, resulting in a relatively small sample size of 468 patients after preprocessing. This sample size is significantly lower than the datasets typically used for training VLMs and LLMs, which often require thousands of samples to achieve robust generalization \cite{vanLargeVisualLanguage2024,tuGeneralistBiomedicalAI2023,fengLargeLanguageModels2023,wangChatCADInteractiveComputerAided2023}. Consequently, the limited sample size affects the generalizability of the model. Another limitation concerns the choice of structured data and imaging modalities. For these preliminary results, only specific cognitive scores and measures were selected, and the study focused on a single imaging modality (MRI), excluding other potentially informative modalities such as PET. These choices in the use of clinical and imaging data may have limited the model’s ability to learn important features, potentially affecting its diagnostic accuracy and completeness. Expanding the model to incorporate additional imaging modalities and a broader range of clinical variables may improve its robustness and clinical relevance. Furthermore, while the use of synthetic reports is an innovative and effective approach for facilitating model learning, the lack of clinical validation for these reports remains a key limitation. Validation by clinical experts is critical for ensuring the quality, accuracy, and real-world relevance of the generated reports.

%\vspace{-0.05in}
Addressing these limitations will be a key focus of future research. We plan to expand the study to other imaging modalities such as PET, including not only the OASIS cohort-4 but also other cohorts from the same dataset and from other datasets such as ADNI and AIBL. This expansion will provide a more comprehensive view of Alzheimer’s pathology and improve the model’s accuracy and generalizability. Future research will also focus on optimizing the model’s performance by improving the integration between VLMs and LLMs, experimenting with new configurations, and optimizing the model’s architecture. Furthermore, collaborations with clinical experts will be sought to validate the generated reports, ensuring greater clinical relevance and contributing to the overall reliability of the proposed methodology. Finally, the current framework will also be explored for other neurodegenerative diseases beyond Alzheimer’s to assess its applicability in different diagnostic contexts. This will require adapting the model to handle different types of neuroimaging data.
\section*{Acknowledgment}

This work has emanated from research supported in part by Taighde Éireann – Research Ireland, formerly Science Foundation Ireland, under Grant Number Research Ireland/12/RC/ 2289\_P2 (Research Ireland Insight Centre for Data Analytics) and co-funded by the European Regional Development Fund in collaboration with the Research Ireland Insight Centre for Data Analytics at Dublin City University. F. Chiumento was also supported by the SFI Centre for Research Training in Machine Learning (ML-Labs) at Dublin City University.
\bibliographystyle{IEEEtran}
\bibliography{references}

\end{document}